# 自然语言处理评测中的问题与对策

董青秀[1,2], 穗志方[1,2], 詹卫东[1,3], 常宝宝[1,2]

（1. 北京大学 计算语言学教育部重点实验室，北京 100871；
2. 北京大学 信息科学技术学院，北京 100871；
3. 北京大学 中文系，北京 100871）

**摘要：**自然语言处理中的评测任务引导和推动着技术、模型和方法上的研究。近年来，新的评测数据集和评测任务不断被提出，与此同时，现有评测暴露的一系列问题也限制了自然语言处理技术的进步。本文从自然语言评测的概念、构成、发展和意义出发，分类综述了主流自然语言评测的任务和特点，进而总结归纳了自然语言处理评测中的问题和成因。最后，本文参照人类语言能力评测规范，提出类人机器语言能力评测的概念，并从信度、难度、效度三个方面提出了一系列类人机器语言能力评测的基本原则和实施设想，并对评测技术的未来发展进行了展望。

# Problems and Countermeasures in Natural Language Processing Evaluation

DONG Qingxiu[1,2], SUI Zhifang[1,2], ZHAN Weidong[1,3], CHANG Baobao[1,2]

(1. MOE Key Laboratory of Computational Linguistics, Peking University, Beijing 100871, China;
2. School of Electronics Engineering and Computer Science, Peking University, Beijing 100871, China;
3.Department of Chinese language and Literature, Peking University, Beijing 100871, China)

**Abstract:** Evaluation in natural language processing guides and promotes research on models and methods. In recent years, new evaluation data sets and evaluation tasks have been continuously proposed. At the same time, a series of problems exposed by existing evaluation have also restricted the progress of natural language processing technology. Starting from the concept, composition, development and meaning of natural language evaluation, this article classifies and summarizes the tasks and characteristics of mainstream natural language evaluation, and then summarizes the problems and causes of natural language processing evaluation. Finally, this article refers to the human language ability evaluation standard, puts forward the concept of human-like machine language ability evaluation, and proposes a series of basic principles and implementation ideas for human-like machine language ability evaluation from the three aspects of reliability, difficulty and validity.

## 0 引言

近年来，随着自然语言处理领域模型和方法的迅捷发展和快速更迭，技术的突破不断引发人们对自然语言处理（Natural Language Processing,

**收稿收稿日期：******-**-**；**定稿日期：******-**-**
**基金项目：**国家科技创新 2030“新一代人工智能”重大项目（2020AAA006701）；国家自然科学基金（U19A2065）

NLP）的进一步探索和评估。因此，伴随着大规模评测数据集的新型评测任务如雨后春笋般涌现，如 GLUE 等 NLP 评测如火如荼地开展起来。

NLP 评测作为对机器理解、处理、应用自然语言能力的一种评估和量化手段，是 NLP 领域的技术水平和研究进展的直观体现，也为相关方向的模型和方法的发展提供了标杆和方向，激励着研究者们更多地参与到相关方向的研究中，是 NLP 相关研究的工具和重要驱动力。

与此同时，Trichelair 等[1]研究者也对部分评测本身的科学性和客观性提出了质疑。研究人员经过设计诊断样例等方法，发现 WSC[2]等多个数据集存在偏差，导致机器模型可能仅仅因为一些与测试意图无关的线索而在评测上取得很好的成绩，实际上机器并没有获得处理相应任务所需要的语言能力。

除了多种类型偏差的存在，评测指标的失真也影响了对机器的语言处理能力的准确评估和比较。BLEU 等[3][4]基于 N-gram 重叠的自动指标在评测中被反复证明与人工评测差距较大，BEER 等[5][6]依赖人工标注的指标则又难以应对不同领域的大量评测需求。

可见，尽管 NLP 评测的数量越来越多，评测的质量和效力却参差不齐。自然语言处理领域亟需对评测本身的质量提出目标、建立一些基本原则，规范 NLP 评测的任务设计，从而保证评测对整个 NLP 领域的发展起到正确的、科学的推动作用。

为此，本文对机器语言能力评测中的问题和对策进行了归纳和探讨。首先，介绍了 NLP 评测的概念、一般构成、发展历程和意义影响。进而，从现有 NLP 评测出发，分类详细介绍了主流的评测。基于对现有评测的调研，本文详细阐述和归纳了 NLP 评测中的问题，探究了导致该类问题的成因。综合上述讨论，本文参照人类语言能力评测规范，提出类人机器语言能力评测的概念，从信度、难度、效度三个不同视角，尝试提出了一系列类人机器语言能力评测的基本原则和实施设想，希望为未来的机器语言能力评测提供相对明确的方向和规范，推动评测向更科学、更有效、更系统的方向发展进步。

## 1 NLP 评测简介

### 1.1 NLP 评测的概念

NLP 评测是指以定义规范的任务为载体，根据机器在任务上的表现评估机器理解、处理、应用自然语言能力的活动。评测需要给出体现机器单方面或者综合能力的量化成绩，为后续自然语言处理研究提供基准和比较标准。

### 1.2 NLP 评测的构成

一般来说，一个完整的NLP评测由核心部件和外围部件共同构成。如图1所示，核心部件指评测任务、评测数据集和评测方法，是评测中相对固定和关键的部分，由评测组织者提供。其中，评测任务是NLP评测的内核，决定了组织评测的目标，即旨在考察机器的何种能力，同时给出了任务的形式化定义，即输入和输出的具体形式、有无辅助信息等。评测数据集与评测任务相辅相成，评测组织者根据评测任务的形式化定义构建或改造数据集，以服务于机器模型的训练和测试，评测数据集也是评测任务的实现载体。为了评测的公平性，需要事先划分训练集、开发集和测试集，近期Matt Gardner等[7]又提出在数据集中增设对比集的想法。在训练集和开发集上训练模型之后，将模型在测试集上得到的结果与测试集预设的参考答案比较评分，一般使用准确率、F1、BLEU等自动指标，近年来Tianyi Zhang等和Thibault Sellam等也分别提出BERTScore[8]、BLEURT[9]等基于预训练的指标，主要用于文本生成质量的评估。

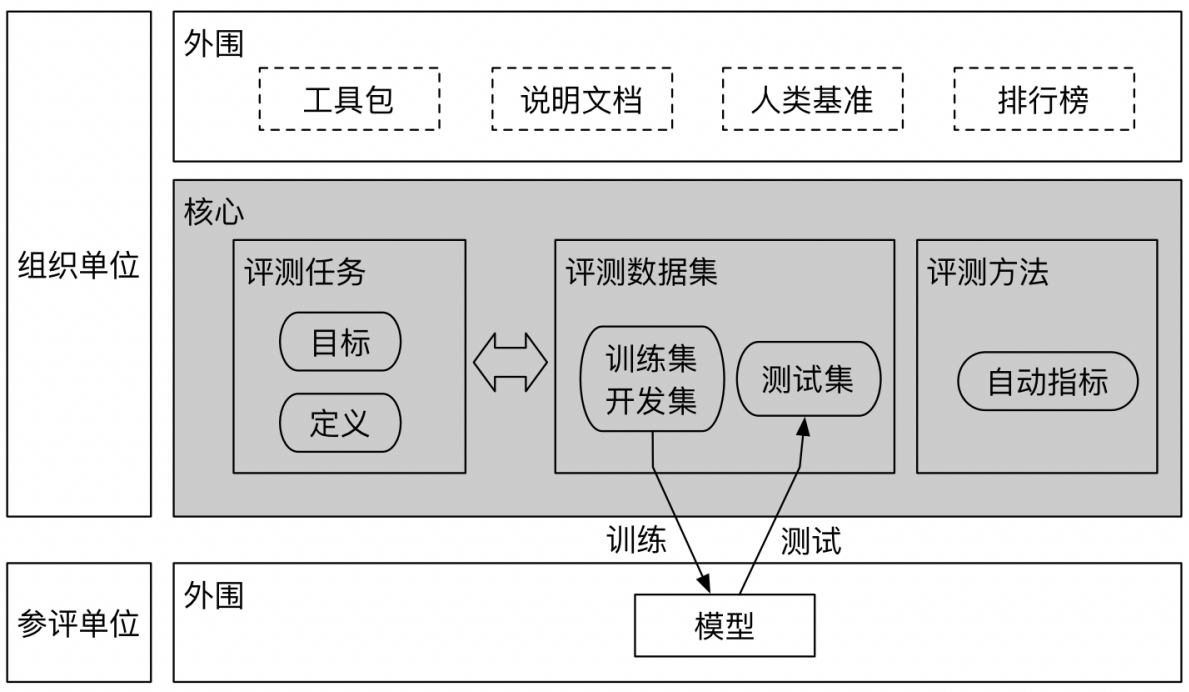

**图 1 NLP 评测的一般构成**

除了上述核心部件，不同评测包括了不同的外围部件。外围部件指容易发生变更、选择性提供的部分，包括由参评单位提供的模型和由组织单位提供的人类基准、排行榜、工具包和说明文档等。以小样本对话语言理解评测FEWJOINT[10]为例，评测组织单位提供了配套小样本工具平台METADIALOG，方便研究者快速开展实验，间接鼓励更多研究者参与评测，从而推动该任务上的研究进展。此外，完整的评测往往会给出人类在该任务上的表现作为参考基准，同时将人类基准和当前较好的模型得分展

示在排行榜上，以便相关研究者及时了解该任务上的最新进展，推动评测的公平性和公开性。

### 1.3 NLP 评测的发展

1950 年提出的图灵测试[11]可视作最早的 NLP 评测方案，自此 NLP 评测进入萌芽期。图灵测试的具体方案是：如果一台机器能够与人类展开对话（通过电传设备）而不能被辨别出其机器身份，那么称这台机器具有智能。然而，图灵测试缺乏量化评估指标，也无法考察机器某项具体能力，故只能作为 NLP 评测的初步构想。

1966 年，ALPAC 报告[12]的发布极大地削减了美国用于机器翻译研究的资金，这也引发了其他国家削减 NLP 研究经费的连锁效应。在后来很长一段时间内，整个 NLP 领域的研究陷入低潮期。在此背景下，NLP 评测也随之进入冷淡期[13]。

1987 年该状况开始改变,NLP 评测进入发展期。这个时期出现了一系列语音处理评测[14]，和文本理解评测[15]。1992 年，美国国防部和美国国家标准技术研究院（NIST）共同发起 TREC[16]，第一个信息检索方向的评测会议应运而生，满足了信息和文档检索研究界的需求。此后，自然语言处理相关会议的数量和研究者的数量不断增长，评测对该领域发展的重要性也越来越被人们认识到，著名的 RTE 系列评测和 SENSEVAL 系列评测都是在这个时期被提出的。

2016 年开始,随着深度学习的发展和应用,NLP 评测进入高涨期。2017 年，Morgenstern 等[2]提出 Winograd Schema Challenge (WSC)数据集，以评测机器解决指代消解和常识推理问题的能力。同年，包含 10 万个问答样例的 SQuAD 数据集[17]公布，推动了此后具有代表性的模型的大量涌现，极大地促进了机器阅读理解领域的发展。除此之外，高涨期还出现了 GLUE 等[18]综合多任务评测基准，目的是覆盖足够大的自然语言处理领域，旨在推动通用的、鲁棒的自然语言理解系统的研究。高涨期的 NLP 评测呈现出 3 个特点：1.评测基准数量多；2.评测数据集规模大；3.评测迭代快。

2018 年以来，虽然新的 NLP 评测如雨后春笋般涌现，NLP 评测依然处于高涨期，但越来越多的研究者开始探究评测中可能存在的偏差并反思评测本身的正确性。2018 年 Trichelair 等人[1]和 2019 年 McCoy 等人[19]提出模型可能仅仅凭借学习高频样例的启发式规则或者问题与答案之间的简单词汇关联而在评测中取得很好的成绩，但可能并没有学到实质的信息，导致模型在下游任务上表现不佳。2020 年，Matt Gardner 等人[7]提出数据集构建者应当在训练集和测试集的基础上增设人工构建的对比集，减小系统偏差对评测结果的影响。随着人们对评测过程中的问题的进一步探究和反思，我们期待未来 NLP 评测能步入稳定发展期，借助于评测基本原则和规范的指导，评测将更加规范化、体系化，不断向类人类语言能力评测逼近，以充分考察模型各方面能力，为 NLP 各个子领域的发展起到引领作用。

### 1.4 NLP 评测的意义

NLP 评测对于自然语言处理的各个方面都有着巨大的指导意义，具体表现在如下 3 点：

1.评测成绩作为机器理解、处理、应用自然语言能力的量化结果，是对模型理解和处理文本能力的直观说明和展示，体现了每个阶段自然语言处理领域的水平和进展；

2.作为 NLP 各子领域的标杆，评测为模型的发展提供了方向，帮助各子领域树立短期和长期优化目标；

3.评测使得模型的能力直观可比，激励着研究者们更多参与相关方向的研究，极大地促进了自然语言处理领域的发展。

正是由于 NLP 评测的重大指导意义，如果评测本身存在问题或偏差，将严重影响相关领域工作价值评定的公平性和客观性，还可能掣肘模型和方法的迭代进步，导致 NLP 研究的方向偏离正确发展的轨道。因此，NLP 评测本身的科学性和合理性的意义也至关重要。

## 2 主流 NLP 评测介绍

NLP 能力评测有多种分类方式，例如根据评测过程中的是否关注模型内部表现而分为白盒评测和黑盒评测，根据评测过程是否有人工参与分为人工评测和自动评测等。由于自然语言处理涉及的种种能力紧耦合，因此很难在一个完整的系统中评测代表所观察功能的一组独立变量，这增加了评测本身的复杂性和公平可靠性。为了尽可能降低复杂性、保证公平可靠，现在主流的评测方法大多是自动评测和黑盒评测。

下文按照考察的能力分类介绍这些主流 NLP 评测的基本情况。

### 2.1 单项能力评测

根据评测中用到数据集数量的不同，单项能力评测又可分为单数据集单项能力评测和多数据集单项能力评测。然而，目前人们尚未将各类评测任务与其考察的机器处理自然语言的能力类型规范化对齐，很多情况下，不严格区分单个数据集和其所对应的评测任务。因此，即使不包含1.2中的所有构件，当单个数据集定义了一个有价值的任务时，便可将其视为单项能力评测。

这类评测的门槛较低，故而数量和类型很多，此处仅介绍几类经典的单项能力评测及其相关的数据集。

2.1.1 命名实体识别——CoNLL2003

CoNLL2003[20]作为2003年CoNLL会议的共享任务被提出，是经典的命名实体识别评测。该任务给定一个句子，要求机器正确识别句子中人名、地名等包含名称的短语。CoNLL2003数据集包括1393篇英语新闻文章和909篇德语新闻文章，英语语料取自路透社收集的共享任务数据集，数据集中标注了4种实体类型。

2.1.2 常识推理——WSC

WSC是多伦多大学计算机科学家Levesque于2011年提出的常识推理测试[21]，该测试试图通过向机器询问特别设计的选择题来考察机器常识推理的能力。任务是给定一对句子，要求机器正确判断问题中某一代词的先行词，是一个二分类问题。由于WSC数据集规模很小，仅包含273个问题，后续规模更大的类似数据集，如DPR、COPA等不断被提出。

2.1.3 文本蕴含推理——MNLI

MNLI由纽约大学于2017年发布[22]，是一个文本蕴含的任务。在给定前提下，需要判断假设是否成立，其中因为MNLI主要特点是集合了许多不同领域风格的文本，因此又分为matched和mismatched两个版本的MNLI数据集，前者指训练集和测试集的数据来源一致，而后者指来源不一致。该任务属于句子对的文本三分类（蕴含，矛盾，中立）问题。

2.1.4 关系抽取——TACRED

TACRED是一个拥有106264条实例的大规模关系抽取数据集[23]，这些数据来自于每年的TAC KBP比赛使用的语料库中的新闻专线和网络文本。在每一年的TAC KBP评测中（2009-2015），100个实体作为查询给出，参与的系统需要为其找到对应的关系和对象实体。工作者使用Mechanical Turk来注释源语料库中包含这些查询实体之一的每个句子。对于每个句子，要求群组工作者标注主体和对象实体跨度以及关系类型。

TACRED中涵盖了TAC KBP比赛中使用的41种关系类型和1个无关类型。

2.1.5 情感分类——SST

SST是斯坦福大学于2013年发布的一个情感分析数据集[24]，主要针对电影评论来做情感分类。因此SST属于单个句子的文本分类任务，输入一个句子，要求输出该句子的情感倾向，即输出“非常积极”、“积极”、“中立”、“消极”或“非常消极”中任一类型。SST包含11,855个句子及相应情感标签，更有挑战性的是，SST同时给出了这些句子的语法分析树中215,154个短语的细粒度情感标签。

2.1.6 文本摘要——DUC

自2001年起，NIST组织发布了DUC系列评测数据集，用于评估机器文本摘要能力。文本摘要任务给定一段长文本，要求机器输出保留其主要信息的短句。DUC系列中最被广泛使用的是DUC2004数据集，其包含500组文档-摘要对，文档平均35.6个词，摘要平均10.4个词。

2.1.7 阅读理解——SQuAD

SQuAD是斯坦福大学于2016年推出的机器阅读理解数据集[17]，是最流行的机器阅读理解评测。任务是给定上下文和问题，要求机器在上下文中抽取可作为答案的片段。SQuAD规模较大，包含十万个问题答案对。

在原来的SQuAD的十万个问题答案对的基础上，SQuAD 2.0[25]中新增了超过五万个新增的、由众包工作者对抗性地设计的无法回答的问题。执行SQuAD 2.0阅读理解任务的模型不仅要能够在问题可回答时给出答案，还要判断哪些问题是阅读文本中没有材料支持的，并拒绝回答这些问题。

2.1.8 对话生成——UDC

Ubuntu对话库UDC由蒙特利尔大学Lowe等人于2015年建立[26]，是目前可用的最大的公共对话数据集。该任务给定一段对话和一句可能的回应，要求判断该回应是不是给定对话的下一句。UDC 1.0版本包含约100万条多轮对话数据，以及超过700万条回复和超过19亿个词。UDC 2.0删除了UDC 1.0中的分词和指代消解等处理，而用特殊的符号对这些信息进行表示。

除了上述介绍的单项能力评测，在自然语言处理的各个子领域都有相应的评测任务，不一而足。带有多重标注或者经过改造的数据集还能被应用于多种不同任务的评测，例如CNN/Daily Mail数据集起初作为大型的有监督问答数据集被提出[27]，后来也被广泛用作生成式文本摘要的基准数据集。得益于自然语言处理社区庞大的语料，这样的单项能力评测类型丰富，设计灵活，能充分评测机器完

成具体任务的能力，例如常识推理能力、语义理解能力。

## 2.2 综合能力评测

区别于单项能力评测，综合能力评测一般都聚合了多个数据集，其中每个数据集自身都可以被视为一个单数据集单项能力评测。

2.2.1 DecaNLP

DecaNLP 由 Salesforce 于 2018 年提出[28]，旨在用问答框架统一各种自然语言处理任务。该评测包含了机器翻译、文本摘要等 10 项任务对应的公开数据集，数据样例被统一转换为问题、上下文和回答的三元组。

2.2.2 GLUE

GLUE[18]是 2019 年由来自纽约大学、华盛顿大学、DeepMind 等机构的研究者创建的英语自然语言理解基准和分析平台，是九种语言理解任务的集合。其设计目的是覆盖足够大的 NLP 领域。只有开发出足够通用的工具，才能在这一基准上表现良好，GLUE 的最终目标是推动通用的、鲁棒的自然语言理解系统的研究。

GLUE 中任务分成三大类：第一大类是分类任务，包括语法错误检测和情感分类，这类任务的输入都是一个序列，输出都是一个类别；第二大类是输入是两个句子，输出是二者的语义是否相似对应；第三大类都是自然语言推理相关的任务，输入前提和假设，希望机器能判断二者的关系是矛盾，蕴含还是无关。

2.2.3 SuperGLUE

由于 BERT 等模型的出现，GLUE 基准在新模型的评估方面日渐乏力，很多模型在 GLUE 的大部分任务上都达到了 90 分以上，GLUE 基准在新模型的评估方面渐渐达到上限。研究者决定将其升级为 SuperGLUE[29]。SuperGLUE 保留了两项 GLUE 任务，还引入了五项难度更大的新任务，增加了对机器共指消解和问答能力的考察，提高了这一测试基准的难度。

截至 2020 年底，SuperGLUE 排行榜上效果最好的模型 T5 已经非常接近人类水平。

2.2.4 CLUE

CLUE[30]即 ChineseGLUE，顾名思义，是中文版本的多任务自然语言理解基准和分析平台。其定位是为更好的服务中文语言理解、任务和产业界，作为通用语言模型测评的补充，通过完善中文语言理解基础设施的方式来促进中文语言模型的发展。CLUE 包含了十个中文语言评测任务，其中很多任务所用的数据集是 GLUE 中的数据集中文版。类似地，2019 年 Le 等人提出了法语上的综合评测基准 FLUE[31]，2020 年 Wilie 等人提出了为印尼语提出了 IndoNLU 评测基准[32]。

2.2.5 LUGE

LUGE 是由百度和中国中文信息学会等于 2020 年 8 月联合推出的中文开源数据集合。针对每个自然语言处理问题，LUGE 中收集和整理了多个开源数据集，进行统一的处理并提供统一的测评方式。其包括了情感分析、阅读理解、开放域对话、文本相似度、语义解析、机器同传、信息抽取 7 类任务，汇集来自 11 所高校和企业的 22 个开源数据集。

2.2.6 XTREME

由于过去的大多数综合能力评测仅限于单种语言，无法在统一标准下评估机器能力，也无法对低资源语言作出相应评测。为了解决这一问题，2020 年 3 月，来自 CMU、谷歌研究院和 DeepMind 的科学家们提出了覆盖四十种语言的大规模多语言多任务基准 XTREME[33]。该评测基准覆盖了 40 种类型不同的语言（跨 12 个语系），并包含了 9 项需要对不同句法或语义层面进行推理的任务，实现了语言多样性、现有任务覆盖以及训练数据可用性的最大化。

2.2.7 XGLUE

无独有偶，2020 年 5 月微软发布了 XGLUE 基准数据集[34]，用于评估跨语言预训练模型在跨语言自然语言理解和生成方面的性能。XGLUE 由 11 种任务组成，涵盖 19 种语言。XGLUE 中每个任务的训练数据都是英语数据，因此要求模型具有强大的零样本跨语言迁移能力，考察模型从特定任务的英语数据中学习并将其学到的知识迁移到其他语言的能力。

与近乎同时期提出的 XTREME 相比，一方面，XGLUE 同时包含跨自然语言理解和生成任务，且除了包含 5 个经典单项能力评测任务，还包含从实际应用场景中选择的 6 个新任务；另一方面，XTREME 覆盖的语言、语系比 XGLUE 更丰富。但总的来说，二者的动机都是评估跨语言预训练模型在跨语言任务上的迁移能力，是非常同质的。

2.2.8 GLGE

GLGE 是 2020 年 11 月由四川大学和微软提出的 NLP 评估基准[35]，旨在填补 GLUE、CLUE 等以分类任务为主的评测在文本生成领域的不足。GLGR 包含 8 个英语生成任务，涵盖了文本摘要、问题生成、问答和对话四项能力。此外，作者按难度将 GLGE 设计为 GLGE-简单，GLGE-中等和 GLGE-困难三类，体现了对难度分级的考量。但 GLGE-中等和 GLGE-困难实际是直接通过对训练数据作筛减得到的，难度分级的设计仍然比较局限。

## 3 NLP 评测中的问题

### 3.1 评测缺乏规范性

由于缺乏评测的基本原则或者要求，提出单项能力评测往往被简化成提出一个新数据集，综合能力评测则大多是单项能力评测数据的简单聚合。故而，现阶段 NLP 评测的准入门槛低，这导致评测数量过多而质量参差不齐。如图 2 所示，仅自然语言推理方向[36]，2005-2020 年就有数十个评测被提出。从图 2 还可以发现，2016 年起，评测的数量和规模都开始大规模增长，即进入了 1.3 所述的评测发展高涨期。

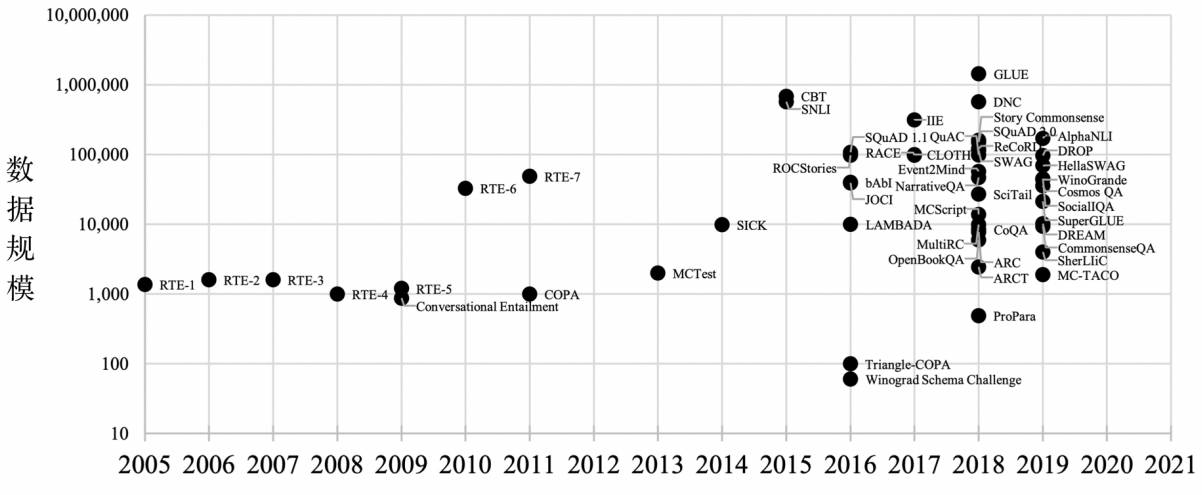

图 2 2005-2020 年提出的自然语言推理（NLI）评测

在评测数量过多的情况下，针对同一项任务的多个评测集给出的结果可能矛盾，但缺乏统一的规范来评估和比较评测本身的合理性。此时，研究者往往各自采用对自己的模型最有利的数据集并声称达到了最好结果，导致后续研究者难以客观比较和超越。

### 3.2 评测效力式微

面对参数量越来越庞大的模型，大部分现有评测难以通过传统指标和单一排行榜明显区分机器模型和人类在测试集上的表现。具体言之，在评测任务提出不久后，机器模型得分往往就已接近甚至超越人类得分。如图 3 所示，我们选取了 12 个提供人类基准和排行榜的评测(具体信息参见表 1)，统计了其提出时最好模型的表现、当前最好模型的表现与人类水平基准。对于这 12 个评测，当前最好模型表现与人类基准的平均差距已不足 3 个百分点。在 SQuAD2.0 等评测上，当前最好模型表现甚至超越了人类基准。但上述现象并不意味着机器语言能力已与人类语言能力不相上下，其真正反映的问题是当前评测在评估机器语言能力方面的效力不足。

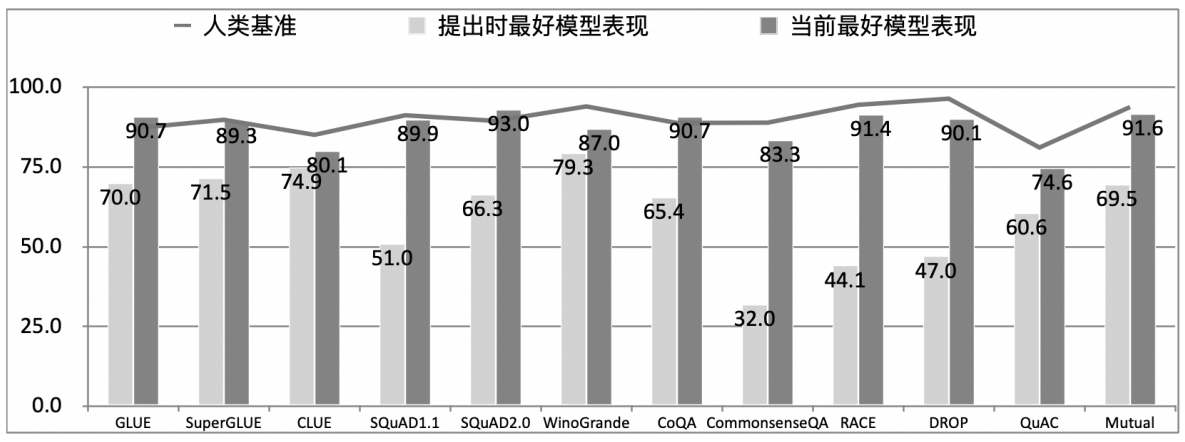

图 3 12 类评测上的人类基准与模型表现

乔姆斯基提出[37]，语言能力是人内在的语言系统的属性，指人具有句法结构系统、词汇系统等语言知识的能力。语言表现是人用语言做事的外在行为特征，是语言能力的外化呈现。语言能力对应着“懂一种语言”（可能属于无意识地知道）；语言表现对应着“能在行为上表现出语言行为”。对比当前最好模型的表现和人类基准可知，现阶段大部分评测上，最好的机器模型表现与人类表现的差距都小于 10 个百分点，在部分评测上甚至超过了人类水平。这说明现阶段大部分 NLP 评测只是在测试语言表现而非语言能力，无法客观反映机器语言能力与人类语言能力的真实差距。

### 3.3 评测数据集偏差

3.3.1 类型分布偏差

同一评测数据集内可能包含多种类型的数据，如果这些类型分布情况存在较大差距，我们称该数据集存在的类型分布偏差较大，这种偏差通常由众包过程引入。一个典型的例子就是机器阅读理解任务中的问题类别分布偏差，在构建问题答案对时，若不加约束地请标注人员拟写问题，可能数据集中大部分问题都是问“是谁……”，这导致评测实际上只是在评测机器阅读和理解人名或指代消解的能力，而无法真实反映机器阅读和理解文本的综合能力。

已经有一系列工作通过人为限制控制类型分布偏差的产生，对于英语问答任务，保持问题类型平衡的一种方法是控制每个问题的第一个单词的分布，参考 CoQA[38]和 CommonsenseQA[39]的创建过程中采用的方法。为了完全规避问题类型偏差，可以在数据标注前设计规范的限定规则，例如等比例随机给众包工作者分配问题类型。

3.3.2 答案分布偏差

答案分布偏差是最容易检测和避免的一种偏差。例如在多项选择题中，如果正确答案以更高的概率出现在某个位置的选项，即答案分布不均匀，模型在学习过程中可能就会引入选项次序信息，尽管模型可能因此在评测中取得了更好的成绩，但显然这并不是我们期待模型具有的语言能力。对于二

分类任务，合理的评测正确答案应该以 50%的概率分布在两个选项中，所以按照服从多数的原则建立的多数类基线应该只有 50 分，但 DNC 评测中的 MegaVeridicality 子任务[40]由于类别标签分布不均而具有 67%的多数类基线，这样的评测数据集存在较大的答案分布偏差。

3.3.3 无关线索偏差

机器并非通过掌握所要评测的语言能力，仅仅利用一些无关的线索推出正确答案的情况，可称为无关线索偏差。

2019 年，T. Niven 和 H. Y. Kao 发现 BERT 在多个评测上取得很好成绩可能只是因为模型学习到了一些虚假相关的统计线索[41]，例如“不”、“是”这种词。他们还通过构造等价对抗数据发现，在对抗数据下 BERT 模型的效果基本等价于随机分类器的效果。这充分说明，无关线索偏误对模型能力评估的误导性，这种情况下，评测本身的效力和信度大大降低了。

无关线索偏差是最容易在众包标注的过程中被引入的偏差类型。McCoy 等在 2019 年提出了统计自然语言推理模型可能利用的三种表层句法偏差（词重叠、子序列重叠、成分重叠），并由此构建了 HANS 诊断数据集[19]。McCoy 等用 HANS 数据集在众包构建的 MNLI 数据集上测试，发现了模型在 MNLI 上成绩“虚高”的问题。图 4 是可分解式注意力模型（DA）、增强顺序推理模型（ESMI）、叠增解析器神经网络（SPINN）和预训练模型（BERT）在 MNLI 测试集上的准确率，图 5 则是各个模型在 HANS 数据集上的准确率。可以发现，各个模型在 MNLI 测试集上呈现出的高分并不意味着模型本身文本蕴含推理能力较强，而是模型利用了 MNLI 数据集上的表层句法偏差“蒙”对了答案，如图 5 所示，模型只要在遇到前提和假设存在词重叠、子序列重叠或成分重叠时直接判断结果为“蕴含”，就能在 MNLI 的测试集上取得高分。

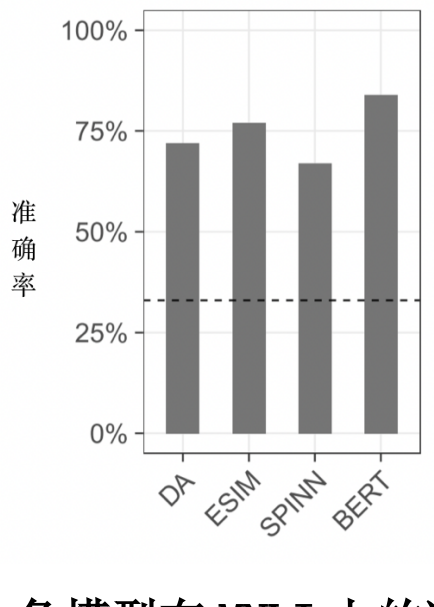

**图 4 各模型在 MNLI 上的准确率**

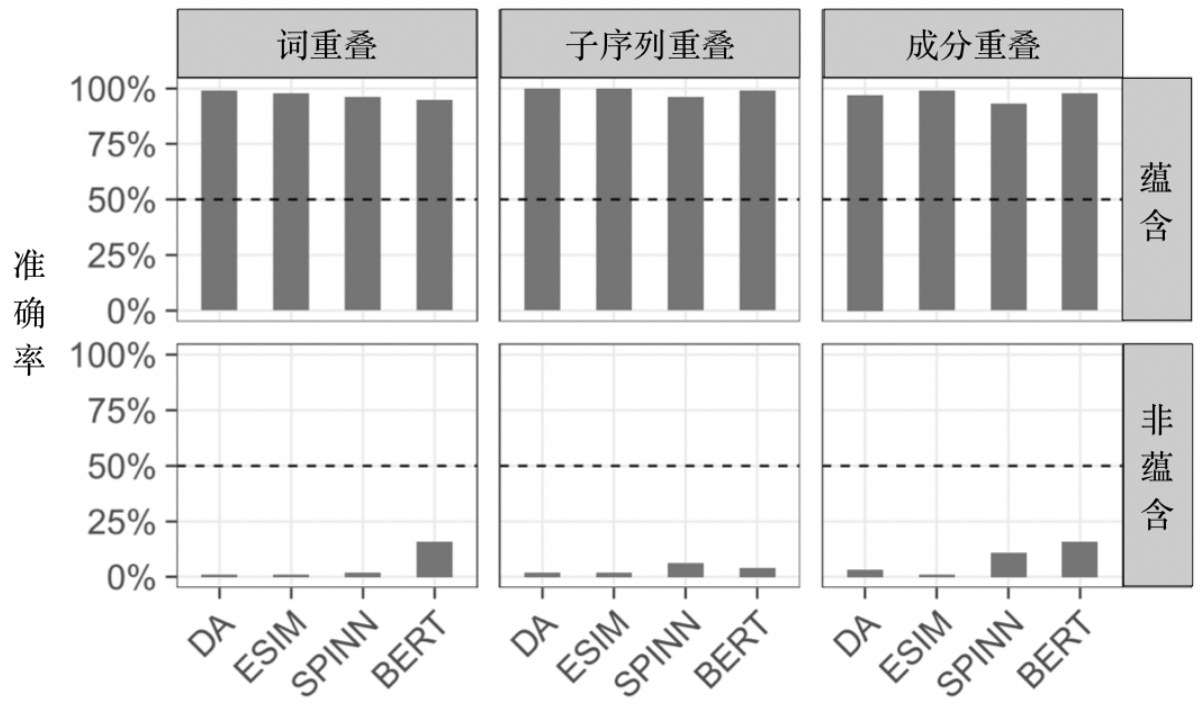

**图 5 各模型在 HANS 数据集上的准确率**

事实上，专家创作的数据也可能存在无关线索偏差[1]。在故事完形评测任务[42]中，专家给故事提供了一个合理和一个不合理的结尾，要求机器选出哪个结尾是合理的。Schwartz 等人的研究发现[43]，仅查看两个的结局即可完成这项测试的准确性高达 72.4%，只需要利用人类写作风格的关联线索偏差来做到这一点，而不需要进行基于事实的推理。例如，他们发现否定性描述通常用于错误的结尾（例如“讨厌”），而正确的结尾更可能使用热情的表述方式（例如“！”）。

3.3.4 关联线索偏差

最难以辨别和规避的偏差是答案和问题的特征之间的偶然相关性引起的关联线索偏差。一个经典的例子是性别偏差，当对有偏的数据进行训练时，在测试集上表现良好的模型可能十分脆弱。Rudinger 等[40]在共指消解中强调了这个问题，他们的研究表明，在性别代名词歧义消除方面，接受过性别偏见数据训练的系统表现更差。

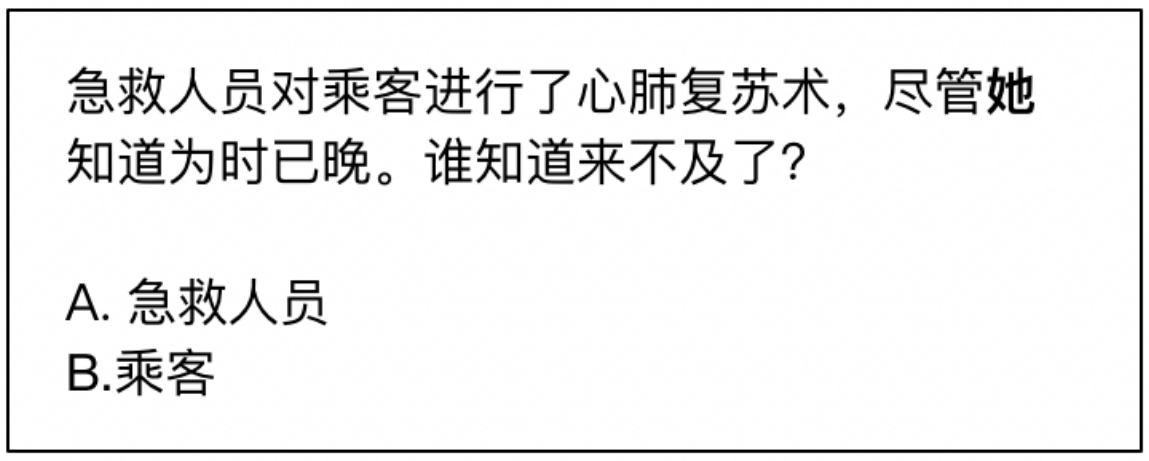

**图 6 Winogender 数据集[44]中的关联线索偏差**

例如，在图 6 的数据集样例中考虑：“急救人员对乘客进行了心肺复苏术，尽管她知道为时已晚。”在确定“她”是谁时，受性别偏差训练数据训练的系统更可能错误地选择“乘客”而不是“急救人员”，因为急救医务工作中，男性性别代词出现在训练数据中的可能性比女性性别代词的高。在来自电影脚本的 Event2Mind 数据[45]中也发现了相

似的性别偏见。此类关联线索偏差往往是源于原始语料分布不均衡。

Sakaguchi 等人[11]的研究发现，众包标注的过程也可能引入相关线索偏差。在编写自然语言数据（例如生成问题或假设）时，众包标注者词汇使用上的习惯或者潜意识也会导致这些关联线索偏差。图 7 是与 WSC 同类的 DPR 数据集[46]中一组有偏示例，事实上，机器只需要根据所问代词“它们”周围的词“食肉”还是“肉多”就可以判断答案是“狮子”还是“斑马”，而不需要对理解整个句子，此类词汇级关联偏差是受众包标注者潜意识影响产生的。

| | |
|---|---|
| 狮子吃斑马因为**它们**食肉。 | **狮子**/斑马 |
| 狮子吃斑马因为**它们**肉多。 | 狮子/**斑马** |

**图 7 DPR 数据集中的关联线索偏差**

### 3.4 评测指标失真

评测指标的选择高度依赖于任务的类型。分类任务的评测指标相对较为统一和明确，如果正确答案或类别标签均匀分布，则分类任务通常使用完全匹配的准确率（Acc）作为评测指标，否则则使用 F 值作为评测指标，某些情况下也可能使用准确率或召回率指标，但 F 值综合考虑了两者，所以近来更广泛地被使用。

文本生成领域缺少这样公认的评测指标，往往容易出现评测指标失真，无法真正反映机器能力的情况。目前主要有两种主要方式对自然语言生成系统进行评测：人工测评和自动化度量。人工测评需要人类标注员在每个模型版本上进行大规模的质量检查，这种方法虽然精度很高但劳动密集型的检查任务十分消耗人力；而像 BLEU[3]和 ROUGE[4]这样自动化测评方法可以对模型进行迅速的评测，但它们仅仅基于 N-gram 重叠对生成文本和参考答案作相似性度量。这些度量标准只对词汇变化敏感，不能识别句子语义或语法的变化。因此，它们被反复证明与人工评估差距较大。诸如此类评测指标使用不当或者指标不统一的情况，也是当前 NLP 评测中暴露且尚未完全解决的重要课题。

### 3.5 评测生命周期短

近几年 NLP 领域的发展进步非常快，这导致部分评测数据集提出后不久，最好的机器模型得分就超过了人类基准，评测系统过快失去效力，这样的评测系统是缺少生命性的。多轮对话推理数据集 MuTual[47]于 2020 年 4 月被提出时，最佳的 RoBERTa 模型在其上的得分仅为 70 分左右，比人类在多轮对话推理任务上的得分低了 20 多分，但不到半年，上海交通大学和华为提出的 MDFN 模型[48]的得分已经和人类基准只有 1-2 分的差距了。

表 1 列出了 3.2 中用到的 12 个提供人类基准和排行榜的评测基本信息，定义提升率为当前最好模型表现与提出时最好模型表现之差与周期的比值（周期为提出时间到当前时间所隔月数）。参考提升率可以发现大部分评测上机器模型的提升率很高，例如 SuperGLUE 或 SQuAD，平均每个月能提升一个百分点，由此可以预测，如果不及时更新迭代，不到一年 SuperGLUE 和 SQuAD2.0 也会彻底走到各自生命周期的尽头。

**表 1 12 类评测的基本信息**

| 评测 | 提出时间[1] | 提出机构 | 任务类型 | 规模 | 提出时的最好模型 | 提出时最好模型表现 | 当前最好模型表现 | 提升率 | 指标 |
|---|---|---|---|---|---|---|---|---|---|
| GLUE | 2018 年 4 月 | 纽约大学等 | 综合 | 9 个任务 | BiLSTM+Att+ELMo | 70.0 | 90.7 | 0.6 | F1 |
| SuperGLUE | 2019 年 7 月 | 纽约大学等 | 综合 | 8 个任务 | BERT+ft | 71.5 | 89.3 | 0.9 | F1 |
| CLUE | 2020 年 11 月 | CLUE 团队 | 综合 | 9 个任务 | RoBERTa-wwm-ext-large | 74.9 | 80.1 | 5.2 | EM、Acc |

[1] 提出时间以论文为准，当前时间仅 SQuAD（官方不再更新排行榜）使用 2019 年 7 月，其余使用 2020 年 12 月

| | | | | | | | | | |
|---|---|---|---|---|---|---|---|---|---|
| SQuAD1.1 | 2016 年 10 月 | 斯坦福大学 | 阅读理解 | 10 万题 | Logistic Regres-sion | 51.0 | 89.9 | 1.1 | F1 |
| SQuAD2.0 | 2018 年 6 月 | 斯坦福大学 | 阅读理解 | 15 万题 | DocQA+ELMo | 66.3 | 93.0 | 0.8 | F1 |
| WinoGrande | 2019 年 11 月 | 艾伦研究所等 | 自然语言推理 | 4.4 万题 | RoBERTa | 79.3 | 87.0 | 0.6 | AUC |
| CoQA | 2019 年 3 月 | 斯坦福大学 | 阅读理解 | 12.7 万题 | Augmt.DrQA | 65.4 | 90.7 | 1.1 | F1 |
| CommonsenseQA | 2019 年 3 月 | 特拉维夫大学等 | 阅读理解 | 1.2 万题 | BiDAF++ | 32.0 | 83.3 | 2.2 | Acc |
| RACE | 2017 年 12 月 | 卡内基梅隆大学 | 阅读理解 | 10 万题 | GA | 44.1 | 91.4 | 1.2 | Acc |
| DROP | 2019 年 4 月 | 加州大学欧文分校等 | 阅读理解 | 9.7 万题 | NAQANet | 47.0 | 90.1 | 2.1 | F1 |
| QuAC | 2018 年 8 月 | 华盛顿大学等 | 阅读理解 | 10 万题 | BiDAF++ | 60.6 | 74.6 | 0.5 | F1 |
| Mutual | 2020 年 4 月 | 浙江大学等 | 对话理解 | 0.9 万题 | RoBERTa | 69.5 | 91.6 | 2.5 | R@1 |

### 3.6 任务缺乏系统性

虽然 GLUE 和 CLUE 等多任务评测声称通用 NLP 评测，试图综合考察模型自然语言理解和处理的能力，但他们本质上并不是完整的、综合的、系统的。各个评测任务都是分立的，只能宏观反映在某个任务上有怎样的成绩，并求一个简单的平均作为最终成绩，这样是不科学的。所谓的综合性评测并不“综合”，只是简单的数据聚合，各个任务没有真正结合成一个系统，没有不重不漏地衡量一个模型的综合能力，无法说明每个任务具体在考察什么，也无法解释任务直接的联系是怎样的。

### 3.7 评测技术单一

现在大部分评测都仅仅依托于一个包含了固定训练集、开发集和测试集的数据集，一方面，一成不变的数据集很容易被机器模型学会、突破，导致评测的效力非常短；另一方面，与训练集基于相同分布的测试集往往只能反映模型在该种分布下的能力，而无法评测模型在真实场景下的性能，无法完成模型对泛化性能的评测。

### 3.8 可解释性较差

完全依赖端到端的方式进行评测存在诸多弊端，例如无法给出带有语言知识的细粒度评测，导致评测的可解释性很差，能给模型改进带来的帮助也十分受限。

## 4 对策：类人机器语言能力评测

针对现阶段 NLP 评测中的一系列问题，我们参照人类语言能力评测方式，提出类人机器语言能力评测的概念。

在评估人类语言能力时，有语言能力综合评测（Language Ability Test，简称 LAT），旨在加强对考生语言综合运用能力的考查，注重知识的联系和融合，增强考试内容的基础性和综合性。LAT 能相对公平高效地实现人类语言水平考核，全面体现考生综合的语言理解和运用能力，保证成绩的有效可信和考试的公平、公正。教育心理学中测试的信度,效度和区分度

借鉴 LAT 的思路，NLP 中的评测也应当是一个有机的整体，充分、客观地反映机器理解和运用语言的能力，即未来面向强人工智能，研究者应当突破传统 NLP 评测的局限，进行类人的机器语言能力评测。

## 5 类人机器语言能力评测的基本原则

机器语言能力评测和人类语言能力评测一样，需要原则指导和规范。教育心理学中使用难度、信度和效度评估试卷，受此启发，我们根据 NLP 评测的现状与问题，从信度、难度、效度着眼，

参照人类语言能力评测规范，提出下列12条类人机器语言能力评测的基本原则。

**表2 类人机器语言能力评测的基本原则**

| 类别 | 原则 | 内容 |
| --- | --- | --- |
| 信度 | 无偏性 | 评测能真实反映出机器解决语言问题的能力，而非机器在特定数据集上的得分 |
| | 鲁棒性 | 相同模型的多次评测结果应当保持一致；评测指标在不同领域或者不同时间跨度使用时漂移尽可能小 |
| | 无悖性 | 评测内部各个评测任务给出的结果不相悖 |
| | 科学性 | 机器语言能力评测给出的结果应和人工评价结果基本一致 |
| 难度 | 挑战性 | 评测需给出更具挑战的任务，使机器模型与人类基准的差距得以显现；难度要在当前机器能力可以触及的范畴 |
| | 区分性 | 评测对当下的各个模型的能力有较强的区分度，排行榜上所有模型结果的离散系数足够大 |
| | 量化性 | 评测需要更具深度的量化，以统一标准为模型内部各部件的作用作出定量评估 |
| | 生命性 | 每个阶段的评测任务都应当符合当时人工智能发展的需要，但也不能过快失效 |
| 效度 | 深层次 | 评测不仅要给出端到端的测试，还要分析模型能否区分相关性和因果性，并对模型处理信息、认知推理等过程的正确性作深层次分析 |
| | 可解释 | 评测要给予未来研究一定的指导，解释起作用的部件并指出模型的缺陷 |
| | 易用性 | 评测应当具有完备的构件（评测数据、任务体系、评测指标、基准模型、人类基准、排行榜、评测报告等），同时具有便捷高效的特性，以便在各环境下高效使用 |
| | 综合性 | 评测任务需要全面多样，同时保证不重不漏，有机结合；最好涵盖跨模态任务 |

## 5.1 信度

信度面向评测的可靠程度，包括无偏性、鲁棒性、无悖性、科学性四项基本原则，是一个机器语言能力评测合理可靠的基本保证。

### 5.1.1 无偏性

无偏性要求评测能真实反映出机器解决语言问题的能力，而非机器在特定数据集上的得分。3.3中介绍了数据集偏差的四种类型，这些偏差妨碍了评测如实反映模型完成某项任务的能力，高信度的机器语言能力评测应对尽量规避。

对于答案分布这类显见的偏差，可以设计一系列简单而通用的基线来检测和消除。例如，用多数类基线(始终选择多数正确答案所在的选项)的结果来检测是否存在答案分布偏差，结果越低说明答案分布越均衡；也可以在构建数据的时候做一些规范化设计，从源头尽量保证无偏性。

### 5.1.2 鲁棒性

鲁棒性包括两个方面，一方面，相同模型的多次评测结果应当保持一致。尤其当设计随机变化的测试集来做评测时，需要保证结果不因为测试集的随机性而发生变动（这要求测试集足够大、足够平衡）。另一方面，评测指标在不同领域或者不同时间跨度使用时漂移尽可能小。

### 5.1.3 无悖性

评测内部各个评测任务给出的结果不相悖。如果有两个任务都能测试模型在某一方面的能力，那么在该评测下，这两项任务测试的结果不应该相悖，否则说明该评测不科学。

一个反例是LUGE评测中，ChnSentiCorp和NLPCC14-SC同为句子级情感分类任务数据集，但在排行榜上出现了在ChnSentiCorp上A模型的得分高于B的，而在NLPCC14-SC上A模型的得

分却低于 B 的情况2。LUGE 系统的排行方式是直接对两个成绩取均值，但二者的结果已经违背了无悖性，这样的结果是不合理的。

5.1.4 科学性

机器语言能力评测给出的结果应和人工评价结果基本一致。检查评测是否科学准确可以采用和人工评价结果相比较的方式，由于机器与人类评价方式的差异，往往无法用相同指标直接作比较，可以考虑用横向对比分析的方法。固定几个经典基线，例如人类基准和 BERT 基准，人工评估这二者在各个任务上给出的结果哪个更好，再看评测给出的各个任务上哪个结果更好，对比二者是否一致来判断评测是否满足科学性。可以使用 Spearman 相关系数来衡量科学性。

## 5.2 难度

难度面向评测的能力，包括挑战性、区分性、量化性、生命性四项基本原则，是一个评测对不同模型区分度的体现。

5.2.1 挑战性

类人机器语言能力评测应当给出更多具有挑战的任务，这体现在现有基线模型得分不会很高，使机器模型与人类基准的差距得以显现；但同时，也要注意这种挑战难度要在当前机器能力可以触及的范畴，如果挑战性过大，评测可能会因为过于超前而暂时无法为研究者提供任何驱动或指导，研究者参与评测的积极性可能会过低。

5.2.2 区分性

评测应当对当下的各个模型的能力有较强的区分度，可以用当前排行榜上所有模型结果的离散系数表示。不同模型的离散系数越大，说明该评测的区分度越大。

5.2.3 量化性

为了衡量结果，现在主流的方法是使用一些评估矩阵作为度量指标，这已经是对机器语言能力一种有效量化。在此基础上，更具深度的量化是类人机器语言能力评测的重要原则，要求以统一标准为模型内部各部件的作用作出定量评估。

5.2.4 生命性

每个阶段的评测任务都应当符合当时人工智能发展的需要，但也不能过快失效。在生命性原则下，至少两年内的最佳模型成绩应低于 90/100，才能保证评测的效力和效果。

进一步，有价值的评测应当在大量模型取得很高成绩（例如 90/100）的时候需要及时迭代更新，不断引领 NLP 发展。

## 5.3 效度

效度面向评测的效力，包括深层次、可解释、易用性、综合性四项基本原则，对类人机器语言能力评测至关重要。

5.3.1 深层次

面向未来的人工智能，我们不应当只用一些表层的、形式化的任务（简单情感分类、抽取式问答）评估模型，而应进一步考查模型深度理解语义的能力。评测内容除了客观信息分析，还要包括主观意图分析、认知过程分析、动态交互分析等。具体来说，从面向弱人工智能的评测到面向强人工智能的评测的一个重要转变是由面向结果到面向过程作评测，评测不仅需要给出端到端的测试，还要分析模型能否区分相关性和因果性，并对模型处理信息、认知推理等过程的正确性作深层次评测。

5.3.2 可解释

类人机器语言能力评测一定要给予未来研究一定的指导。在评测报告（可以自动生成）中应当指出模型的缺陷是什么，面对哪些问题处理得较差或者模型相较于其他模型在理解哪些的能力更强，或者是模型的哪个部件起到了相应作用。

5.3.3 易用性

评测应当具有完备的构件（评测数据、任务体系、评测指标、基准模型、人类基准、排行榜、评测报告），同时具有便捷高效的特性，以便在各环境下方便使用。

5.3.4 综合性

大量模型在部分易解的自然语言处理问题上取得了很好的结果，但是在一些复杂难解的问题上还有很长的路要走。这并不意味着评测可以直接丢弃那些容易的任务，而应从人类语言不同能力（阅读、推理、创造等）出发，综合地评估机器模型。

一方面，评测任务需要全面多样。综合的 NLP 评测既要考查到模型理解文本的能力，也要考查其生成文本的能力；既要涉及命名实体识别等经典任务，也要前瞻性地针对多模态任务和知识驱动任务作设计；既要有面向有监督方法的测试和

[2] Alvin0310 的团队模型和基线 ERNIE 的得分

比较方案，也要为小样本、零样本、无监督等方法提供测试和比较方案。

另一方面，评测需要是一个有机系统。评测设计者应当为评测本身提供充分的可解释性（每个任务评测了模型哪些方面的能力，任务之间的联系与区别，评测结果的解读）。任务与任务之间应当做到不重不漏，有机结合，作为对机器语言能力的完整评测呈现。

值得一提的是，多任务的综合评测最好能够涵盖跨模态任务。跨模态的任务需要模型具有建模和联合不同模态信息的能力，是更加符合人类智能的综合型任务。人类往往在一个多模态场景中理解语言，也能将文本作不同模态的输出。图片、影音信息数据可以作为文本任务的辅助信息，也可以作为生成目标来与自然语言处理任务结合，这是当前人工智能发展的重要趋势。因此，类人机器语言能力评测不应当局限于纯文本任务，需要评估模型对这类多模态信息处理能力。

## 6 类人机器语言能力评测展望

### 6.1 更系统的评测大纲

基于可解释性原则和综合性原则的要求，类人机器语言能力评测需要在体系化的组织和指导下展开。因此，需要事先建立评测大纲，系统梳理 NLP 的各项核心技术、各项评测任务之间的关联性和有机性，建立机器语言能力与人类语言能力相对清晰的对应关系，全面盘点目前 NLP 各项技术的进程。

评测大纲可有效规范和指导未来的评测，避免评测数量过多而质量参差不齐、可解释性差等乱象。

### 6.2 更具挑战的评测任务

类人机器语言能力评测应当以人类语言能力为参照，根据当前机器语言能力发展状况，提出有挑战性的任务。例如心理学上通过语用测试来考察人的语言认知能力，要求被试回答“如果他(她)给你弄的吃的和喝的都不是你喜欢的，你会怎么让他(她)知道？”机器语言能力评测也应当设计此类有深度、考察认知的任务，而非一味设计只需要捕获简单统计规律即可解决的任务。

### 6.3 更科学的评测方法

6.3.1 模组化评测

从任务组织模式的角度来说，可以把某一类标注数据作为一种部件，按照评测需求快速搭建任务框架，实现评测模组化。

例如，把一组未标注的新闻领域语段作为一个一级部件 A1，将关于这组数据的命名实体标注作为一个二级部件 B1，将这组数据的摘要标注作为一个二级部件 B2，将这组数据相关的图片也作为一个二级部件 B3。A1+B2 可以作为一个文本摘要任务，A1+B2+B3 可以作为一个跨模态摘要任务。该种模组化的评测组织模式提高了数据重用率，同时提升了评测的可解释性和综合性，实现了额外特征标注信息和机器模型本身的分离，能促进关于机器语言能力的公平比较和评析。

6.3.2 层次化评测

层次化的评测要求多个子任务之间呈现层级递进的关系，从而体现机器语言能力评测的基本原则中关于难度的各个原则。如图 8 所示，对于包含图文的多模态任务，可以设计如下四种评测层级。各上级子任务对其下级子任务存在包含关系，即模型必须在下级任务上取得较好的表现时，才能完成上级任务，否则模型可能仅仅是因为学到了数据集偏差，例如依赖一些错误的表层线索推出答案。

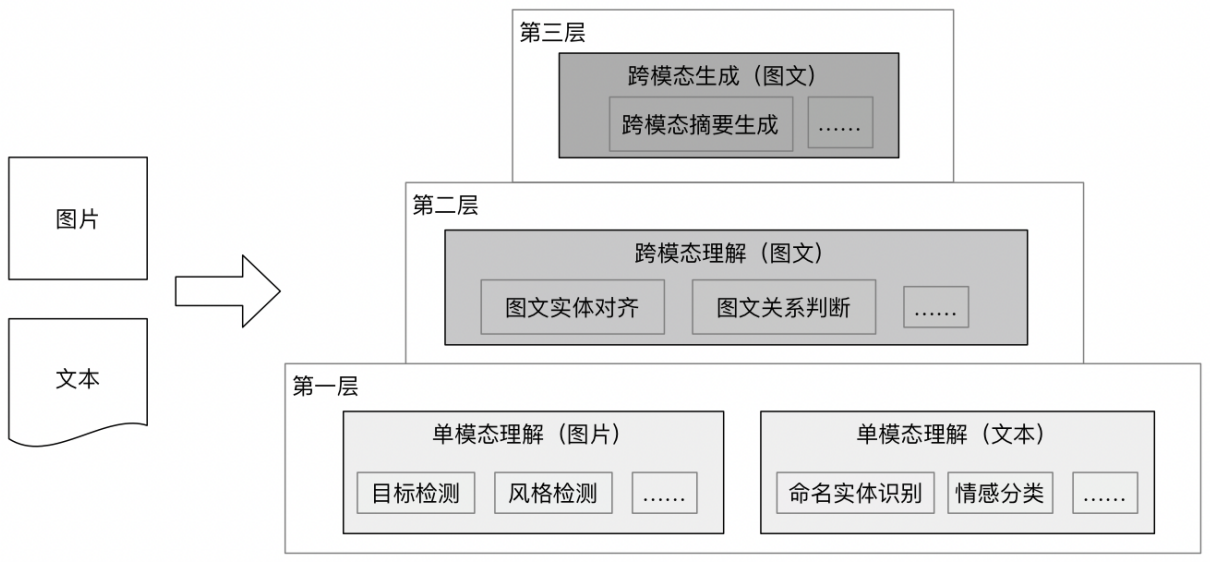

**图 8 层次化评测示意图**

层次化的评测能有效规避评测中的偏差，提供更有说服力评测结果。

6.3.3 交互式评测

人类面试过程中，面试官可能一步步提供信息引导面试者给出更好的答案，整个过程中面试者与面试官是动态交互的状态。借鉴这种思路，机器语言能力评测也可以设计成动态交互式的评测，分角度引导机器思考并一步一步给出目标信息。

6.3.4 动态调整式评测

如前文所述，机器语言能力评测应当满足生命性原则，即评测在至少两年内应当保证评测效力。现阶段，很多 NLP 评测的生命周期短的一大

原因就是采用了固定测试集。尽管模型在训练阶段无法看到测试集数据，但模型不断迭代比较的过程都直接参照测试集的成绩，从长期来看，测试集实际以一种“软开发集”的形式被利用了起来。这种情况下，机器模型在测试集上的表现自然会以较快的速度提升，从而间接导致评测本身生命性的减弱。为了缓解这种情况，增强机器语言能力评测的生命性，可以利用数据增强、对抗攻击等技术，在评测过程中引入随机变动的测试集，增强机器语言能力评测的生命性。

6.3.5 应用检验式评测

机器语言能力评测原则中的无偏性，换言之，是要求评测把模型的泛化性能纳入考量。传统 NLP 评测往往把模型在与训练集呈现相同分布的测试集上的表现作为最终结果，这导致了模型在实际应用场景测试时可能会出现巨大落差。未来，依托于工业界的测试场景和真实环境，研究者可以建立应用检验式的评测，把模型在真实场景上的表现作为评测结果，或者作为辅助信息为评测结果提供参考。

### 6.4 更一体化的评测平台

机器语言能力评测可以建设成通用平台，走一体化的道路。一方面，可以实现评测和诊断的一体化，即通过一系列探针任务把评测和诊断合二为一，自动生成可解释的评测结果和细粒度的诊断报告。另一方面，也可以实现领域和任务的一体化，即把领域内和领域外迁移学习评测，大样本、少样本和零样本评测在相同任务下同时实现。我们期待未来能建立通用的一体化评测平台，组织通用的、可交互可解释的评测与诊断服务，为机器语言能力的研究探索提供高效、准确和全面的反馈。

## 7 总结

本文介绍了现阶段主流自然语言评测，并归纳现阶段自然语言评测中存在的问题为四个类型——数据集偏差、评测指标失真、评测任务不科学和评测技术单一，分别对每类问题的成因进行了深入探析。在此基础上，本文结合目前最新的研究进展与研究思路和成果，针对上述关键问题，提出类比人类语言能力评测的机器语言能力评测，并从信度、难度、效度三个方面设计了共 12 条类人机器语言能力评测的基本原则，希望为未来机器语言能力评测规范的设计提供参考。

目前，自然语言评测虽然处于高涨期，但其在各类任务上暴露的种种问题表明，现阶段的自然语言评测缺少原则和规范的约束，这也导致自然语言处理评测迟迟无法向稳定发展的阶段迈进。未来，自然语言评测应当以具有综合考察能力的类人机器语言能力评测为目标，在参考本文提出的基本原则和未来展望的基础上，采取更多样、更鲁棒的评测手段，科学高效地为机器模型提供客观、公平、类人的评测结果，继续引领和推动自然语言处理领域各类模型、方法的提出和创新。

## 参考文献

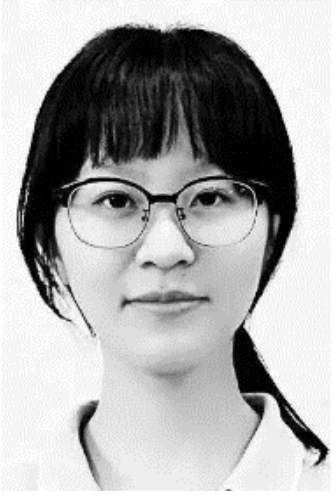

董青秀（1998—），博士研究生，主要研究领域为自然语言处理。

E-mail：qingxiudong@icloud.com

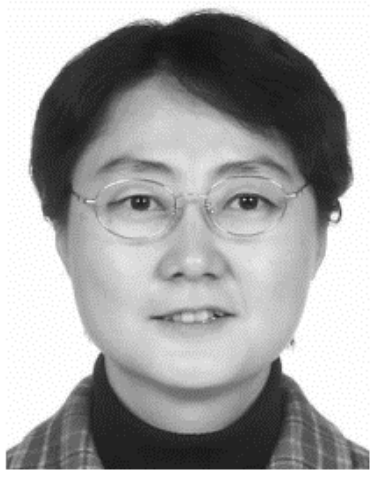

穗志方（1970—），通讯作者，博士，教授，主要研究领域为计算语言学、知识工程。

E-mail：szf@pku.edu.cn

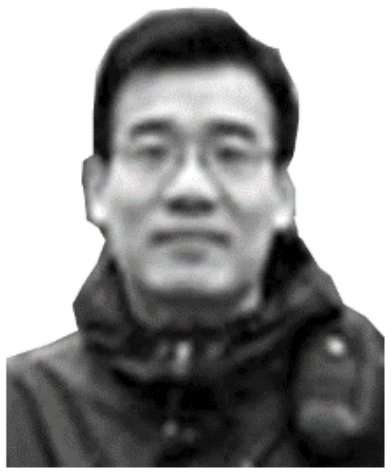

詹卫东（1972—），博士，教授，主要研究领域为现代汉语形式语法、中文信息处理、汉语语言知识工程。

E-mail：zwd@pku.edu.cn